\documentclass[letterpaper, 10 pt, conference]{ieeeconf}

\IEEEoverridecommandlockouts
\usepackage[tracking=true]{microtype}
\usepackage{amsmath} 
\usepackage{amssymb}  
\usepackage{graphicx}
\usepackage{epsfig} 
\usepackage{mathptmx} 
\DeclareMathAlphabet{\mathcal}{OMS}{cmsy}{m}{n} 
\usepackage{times} 
\usepackage{amsmath} 
\usepackage{amssymb}  
\usepackage{xspace}
\usepackage{tabularx}
\usepackage{multirow}
\usepackage{adjustbox}
\usepackage[frozencache,cachedir=minted]{minted}
\setminted{fontsize=\footnotesize}
\usepackage{url}
\usepackage{float}
\usepackage[dvipsnames]{xcolor}

\usepackage[font={footnotesize}]{caption}
\usepackage{booktabs} 
\usepackage{tikz}
\usetikzlibrary{positioning,shapes.geometric,arrows,fit,shapes.symbols,svg.path,tikzmark,calc}
\usepackage{booktabs}
\usepackage{textcomp}
\usepackage{listings}
\usepackage{subcaption}
\pgfdeclarelayer{background}
\pgfdeclarelayer{foreground}
\pgfsetlayers{background,main,foreground}
\usepackage{siunitx}
\usepackage[short]{optidef}
\let\labelindent\relax
\usepackage[inline]{enumitem}

\DeclareMathOperator*{\argmin}{argmin}

\newcommand{\AlgName}{BOMP}
\newcommand{\OptName}{\AlgName{}}

\title{\LARGE \bf
BOMP: Bin-Optimized Motion Planning
}

\author{Zachary Tam$^{1}$, Karthik Dharmarajan$^{1}$, Tianshuang Qiu$^{1}$, Yahav Avigal$^{1}$, Jeffrey Ichnowski$^{2}$, Ken Goldberg$^{1}$
\thanks{$^{1}$The AUTOLab at UC Berkeley (automation.berkeley.edu)}%
\thanks{$^{2}$Carnegie Mellon University, Pittsburgh, PA, USA}%
}

\begin{document}

\maketitle
\thispagestyle{empty}
\pagestyle{empty}

\begin{abstract}
In logistics, the ability to quickly compute and execute pick-and-place motions from bins is critical to increasing productivity.
We present Bin-Optimized Motion Planning (\AlgName{}), a motion planning framework
that plans arm motions for a six-axis industrial robot with a long-nosed suction tool to remove boxes from deep bins.
BOMP considers robot arm kinematics, actuation limits,
the dimensions of a grasped box, and a varying height map of a bin environment to rapidly generate time-optimized, jerk-limited, and collision-free trajectories.
The optimization is warm-started using a deep neural network trained offline in simulation with 25,000 scenes and corresponding trajectories. 
Experiments with 96 simulated and 15 physical environments suggest that \AlgName{} generates collision-free trajectories that are up to 58\,\% faster than baseline sampling-based planners and up to 36\,\% faster than an industry-standard Up-Over-Down algorithm, which has an extremely low 15\,\% success rate in this context. \AlgName{} also generates jerk-limited trajectories while baselines do not.
Website: \url{https://sites.google.com/berkeley.edu/bomp}.



\end{abstract}
\section{Introduction}

Robots are increasingly used for package handling and picking in logistics settings. When transporting thousands of packages each day, reductions in cycle time can significantly increase robot productivity.

Particularly when working in deep bins, package handling robots are often equipped with long-nosed suction tools (e.g., the ``bluction'' tool from Huang, et al.~\cite{huang2022bluction}) to enable them to reach and manipulate packages throughout the deep bin. The long-nosed suction tool also allows the robot wrist and arm to remain far from obstacles and potential collisions.

In deep cluttered bins, contents can shift after each pick, necessitating a strategy to rapidly compute pick-and-place motions using the latest sensor data (e.g., color and depth camera images).

Practical approaches include heuristic planning, optimization-based motion planning, and sampling-based motion planning. A common heuristic trajectory, Up-Over-Down, lifts the package to clear all obstacles, moves horizontally over obstacles to the target location, then lowers. This is easy to implement and has negligible compute time; however, the motion is longer than necessary and often fails when using a long-nosed suction tool and planning in a deep bin. In particular, \emph{vertically} lifting from the bottom to the top of the bin may be kinematically infeasible in deep bin environments.

Optimization-based methods formulate and solve an optimization problem to find the best or fastest trajectory that avoids collisions. Sampling-based methods randomly sample and connect collision-free waypoints to find a path.
The latter two methods yield significantly more successful and faster motions than Up-Over-Down, but at the expense of longer compute times.

\begin{figure}[t]
\centering
\includegraphics[width=1.0\linewidth]{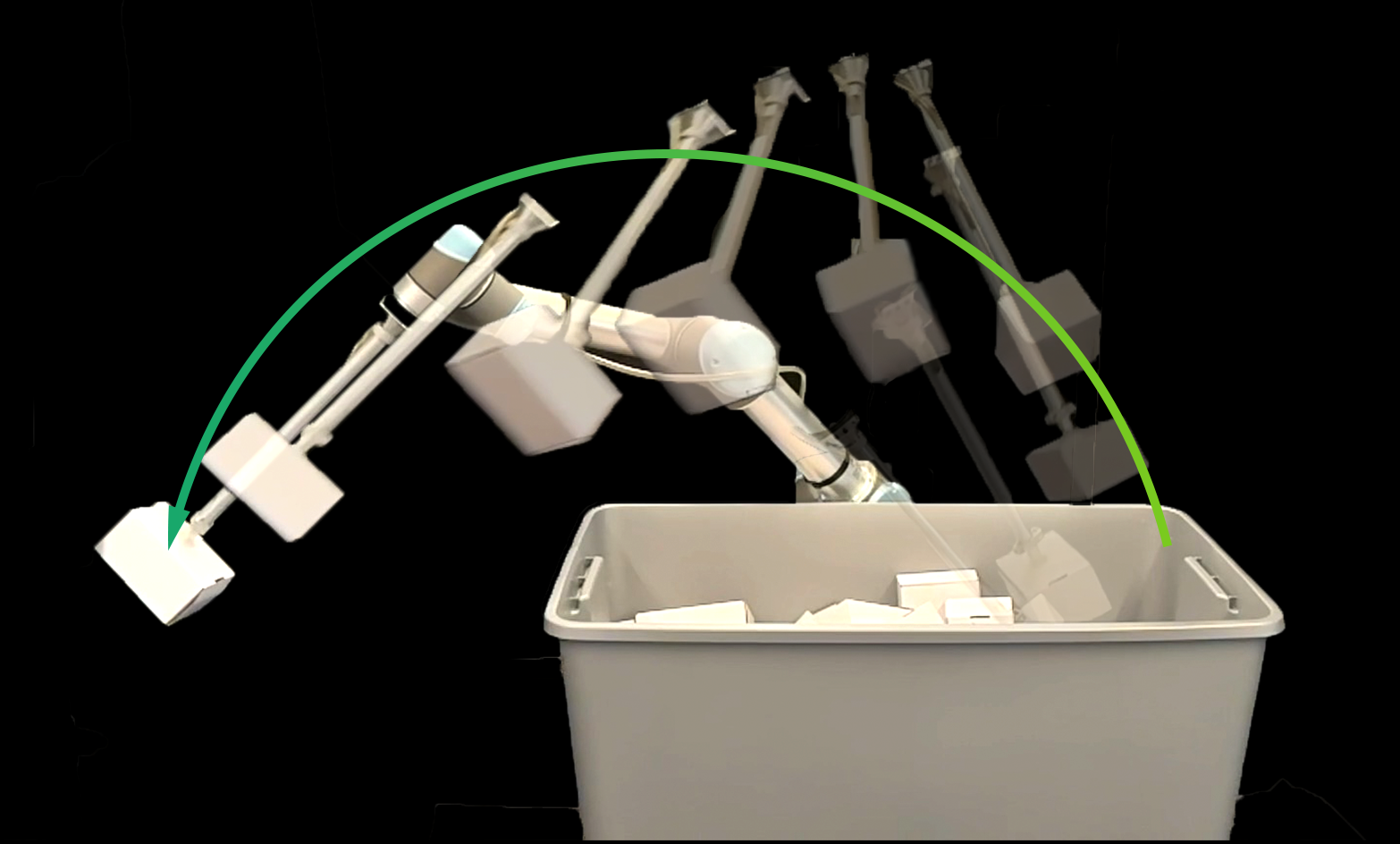}
\caption{\textbf{Bin-optimized motion planning.}
\AlgName{} executing a time-optimized, jerk-limited, collision-free trajectory moving a box from a bin to a drop-off point. We use the long-nosed ``bluction'' tool from Huang, et al. \cite{huang2022bluction} to enable the robot to reach all parts of the deep bin, and an overhead RGBD camera to detect obstacles and target boxes. \AlgName{} uses an optimization-based motion planner to compute the pick-and-place trajectory. In order to speed up the computation, a neural network warm-starts the optimizer. It accepts the obstacle environment, grasped box, and pick and place poses as input, and outputs an initial trajectory.
}
\label{fig:figure_one}
\end{figure}


In prior work, we presented GOMP~\cite{ichnowski2020gomp}, an optimization-based motion planner incorporating time-optimization, obstacle avoidance, and grasp-optimization.  GOMP computes fast motions between pick and place poses, and the grasp-optimization further speeds up motions by allowing pick and place poses to be optimized while retaining the same parallel-jaw grasp contact points.
Subsequently, DJ-GOMP~\cite{ichnowski2020djgomp} further reduces compute time by using
a neural network to warm-start motion planning for time-optimized and jerk-limited trajectories. DJ-GOMP is trained over a distribution of start and end points, assuming a fixed collision environment. 
However, in warehouse settings, boxes often move between picking actions. 

To address changing obstacle environments, we propose Bin-Optimized Motion Planning (\AlgName{}). \AlgName{} finds a time-optimized trajectory while considering collisions between the robot, a grasped box, and the obstacle environment. We integrate \AlgName{} into an end-to-end bin-picking pipeline, which takes as input an RGBD image of the bin and outputs time-optimal trajectories (Figure \ref{fig:figure_one}). 
\AlgName{} modifies and extends DJ-GOMP by adding the dimensions of a grasped box and a height map of the environment as inputs to the warm-start neural network, which enables adapting the trajectory based on environment changes.  The warm-start neural network is trained to handle a relevant distribution of varying collision obstacles, and predicts a trajectory to warm-start the jerk-limited and time-optimizing trajectory optimization. 
We generate a dataset of synthetic collision environments in simulation to train the deep neural network. 

This paper makes 4 contributions:
\begin{enumerate}
    \item \AlgName{}, a time-optimizing, jerk-limited
    motion planning algorithm for picking boxes from deep bins where boxes and obstacles are detected via a depth camera. In experiments, \AlgName{} generates up to 36\,\% faster trajectories compared to an Up-Over-Down baseline implemented with an optimal time-parameterizer~\cite{kunz2012time}, as well as up to 58\,\% faster trajectories compared to Motion Planning Templates~\cite{ichnowski2019mpt} (MPT) which implements a parallelized sampling-based motion planner (PRRT*). \AlgName{} successfully generates collision-free trajectories at a 79\% rate, which is similar to MPT, and significantly higher than the Up-Over-Down baseline.
    \item An end-to-end bin-picking pipeline that uses \AlgName{} to iteratively remove the boxes.
    
    \item A deep neural network trained on 25,000 trajectories generated from simulated scenes to accept a height map, grasped box parameters, a number of trajectory segments, and trajectory endpoints as inputs and to predict an initial trajectory to warm-start the planning.
    \item Data from 96 experiments in simulated environments and 15 experiments in physical environments.
\end{enumerate}

\section{Related Work}

Optimization-based motion planners such as CHOMP~\cite{ratliff2009chomp}, STOMP~\cite{kalakrishnan2011stomp} and TrajOpt~\cite{schulman2013finding} compute motion plans by locally optimizing a trajectory while penalizing collisions or placing barrier functions on collisions~\cite{kuntz2017fast}.  Marcucci et al.~\cite{marcucci2022motion} take a different approach: they decompose the collision-free space into convex regions and use convex optimization to find a collision-free path. Natarajan, et al.~\cite{natarajan2024implicit} use implicit graph search methods to accelerate this path-planning.
GOMP~\cite{ichnowski2020gomp} builds on prior formulations by taking the mechanical limits of the robot arm and the dynamics between waypoints into consideration. It also optimizes over a rotational degree of freedom about the parallel-jaw gripper contact points in each of the pick and placement frames.
DJ-GOMP~\cite{ichnowski2020deep} further extends GOMP by minimizing jerk to reduce joint wear while also significantly reducing computation time by warm-starting the optimization with the output of a deep neural network. 

Prior work has considered model distillation (i.e., one model being trained on the output of one or more different models) for seeding optimization-based motion planners and for achieving more general-purpose models.
In many cases, training an ensemble of models improves prediction performance, but is computationally expensive. However, the ensemble can often be effectively distilled into a compact network~\cite{hinton2015distilling,bucilua2006model}. 
DJ-GOMP used model distillation to improve GOMP's running time.
While the repeated optimization executed in GOMP could take up to several minutes to finish, a forward pass in a compact neural network can be executed in milliseconds. 
DJ-GOMP exploits this feature of neural networks to compute similar robot trajectories faster for a known collision environment. However, when the collision environment is variable, and only known during runtime, the compute time of DJ-GOMP increases by several orders of magnitude since the neural network warm-start is no longer valid. \AlgName{} addresses this by learning a representation of the collision environment and the grasped box, in addition to robot trajectories.  

Warm-starting an optimizer with a near-optimal solution can significantly increase the solver's performance while greatly reducing the number of iterations required to reach sufficient optimality~\cite{mordatch2014combining}.
In reinforcement learning, learning a new task can be warm-started by transferring features from old tasks the agent has already mastered~\cite{taylor2009transfer}. Memory of motion~\cite{lembono2020learning} is another method that uses an offline learned policy to warm-start a control solver, and was shown to reduce the computation time in locomotion problems, and to increase the performance of nonlinear predictive controllers~\cite{mansard2018using}. In GOMP~\cite{ichnowski2020gomp}, compute speed is limited by the number of iterations required to find the optimal trajectory duration. To address this, DJ-GOMP~\cite{ichnowski2020djgomp} uses a neural network's output to warm-start the optimization with an approximation of the optimal trajectory, ultimately resulting in faster convergence. \AlgName{} uses a similar approach, taking into consideration the grasped box and the current state of the collision environment to predict the optimal collision-free trajectory.

In recent years, researchers have explored the approach of bypassing the optimization step and using purely learning-based methods for motion planning.
Motion planning algorithms can require complex cost functions, and learning-based methods, such as learning from demonstrations (e.g., \cite{ye2017guided,abbeel2008apprenticeship,rana2017towards}) can reduce the amount of hand engineering required.
Some learning methods (e.g., ~\cite{huang2024neural, das2024motion}) focus on increasing the sampling efficiency in sample-based methods, for example by using non-uniform sampling~\cite{ichter2018learning} or reinforcement learning~\cite{zucker2008adaptive}
To address learning in a complex obstacle environment, Qureshi et al.~\cite{qureshi2019motion} encode a point cloud of the obstacles into a latent space and use a feed-forward neural network to predict the robot configuration at the next time step given an initial state, goal state, and the obstacles encoding. However, learning-based methods tend to generate less optimal trajectories and often fail to generalize to new environments.

Although pick-and-place tasks are readily addressed by sampling-based motion planners~\cite{schmitt2017sampling, ichnowski2014prrt},
and despite the advances made via learning-based methods, the non-negligible convergence rate of these planners in high dimensions prevents them from performing well in picks-per-hour.

In~\cite{coleman2014thunder, berenson2012lightning}, the authors use past experiences to reduce this convergence rate. However, these methods are most effective when the obstacle environment remains mostly the same between executions, whereas the bin environment changes with every pick.

To detect target boxes and the obstacle environment, \AlgName{} uses both a depth image and RGB segmentation data computed using the Segment Anything Model ~\cite{kirillov2023segany}.
Previous methods such as Dex-Net \cite{mahler2017dex} use only point clouds. The segmentation masks generated by SAM~\cite{kirillov2023segany} and successor models such as FastSAM~\cite{zhao2023fast} can provide valuable information on which points belong to the same object.

\section{Problem Statement}
\label{sec:problem}
We consider a set of $n$ rigid rectilinear boxes selected from a finite set of known candidate dimensions and randomly placed into a deep-walled bin (where the bin depth is greater than twice the largest box dimension). Given an overhead RGBD image of the bin before each pick, we extract boxes using a six-axis industrial robot arm with a long-nosed suction tool. Because of the long-nosed suction tool, we assume that the robot wrist and upper arm will never collide with the environment. We further assume that the bin is in a fixed, known pose and that the obstacle environment is fixed during each robot motion (thus, we run open-loop motions). An example problem setup is shown in Figure~\ref{fig:figure_one}.

We consider the problem of transporting boxes from the bin to a designated dropoff location. This problem is composed of two subproblems that must be solved repeatedly for each box:
(1) detecting and selecting a box and a suction point; and
(2) computing a fast and collision-free trajectory to transport the box from that grasp pose to the dropoff point.

Let $q \in \mathcal{C}$ be the complete specification of the degrees of freedom, or \emph{configuration}, of a robot, where $\mathcal{C}$ is the configuration space.  Let $\mathcal{C}_\mathrm{obs}$ be the set of robot configurations in which the robot is in collision with the environment, and let $\mathcal{C}_\mathrm{free} = \mathcal{C} \setminus \mathcal{C}_\mathrm{obs}$ be the free space. Let $x_t$ define a state composed of $q_t$, $\dot q_t$, and $\ddot q_t$. Let $\tau = (x_0, \ldots, x_H)$ be a trajectory composed of $H+1$ robot states, where each state is separated in time by $t_\mathrm{step}$. Let $Q \subset \mathcal{C}$ be the kinematic limits of the robot, and let $\dot Q$, $\ddot Q$, and $\dddot Q$ be the velocity, acceleration, and jerk limits.

Before each pick, an overhead RGBD camera provides an RGB image $\mathbf{y}_\mathrm{RGB} \in \mathbb{R}^{h\times w\times 3}$ and a corresponding depth image $\mathbf{y}_\mathrm{D} \in \mathbb{R}^{h\times w}$ of the bin scene.

The objective of subproblem (1) is to compute $q_0$ from $\mathbf{y}_\mathrm{RGB}$ and $\mathbf{y}_\mathrm{D}$. The objective of subproblem (2) is to compute $\tau$ such that it picks the box at $q_0$ and moves it to the target location, and $q_t \in \mathcal{C}_\mathrm{free} \cap Q$, $\dot q_t \in \dot Q$, $\ddot q_t \in \ddot Q$, and $\dddot q_t \in \dddot Q$ for all $t \in [0, H]$, while minimizing motion time. 
\section{Method} \label{iv-methods}

\subsection{Grasped Box Shape Estimation} \label{sec:sam}

To select a target box, we find the highest point among the boxes in the bin, then prompt SAM \cite{kirillov2023segany} with the corresponding  pixel in $\mathbf{y}_\mathrm{RGB}$.
This method heuristically picks less occluded boxes and empties the bin in a top-down order.



To determine the target box dimensions, we apply the SAM output 
mask to $\mathbf{y}_\mathrm{D}$ and create a segmented pointcloud.
We fit a rectilinear box to this segmented pointcloud using RANSAC~\cite{fischler1981ransac}, which returns up to 3 orthogonal planes. For each plane, we compute the bounding rectangle of the inliers.
We compute potential matches by comparing the face areas and edge lengths of the observed rectangles against each option in the finite set of known candidate dimensions (Section~\ref{sec:problem}). We aggregate results using both metrics based on the smallest difference with any potential match.
In the case of a tie, we conservatively assume the largest possible box size. 
Using this estimate of the box dimensions, we estimate the box vertex positions. 


\subsection{Suction Grasp Selection}
To determine the initial grasp configuration $q_0$, we consider suction points at the center of each of the visible faces of the target box. We sort these grasp candidates by their normals' dot product with the positive z-axis (i.e., how close they are to a top-down grasp, as these give the robot the most freedom to move). We then evaluate them in order, rejecting those where no IK solution exists or where the robot would be in collision with the environment. We terminate this search as soon as we find a reachable collision-free grasp.

The UR5 robot used in our experiments has parallel revolute joints, thus it generally has multiple IK solutions for a given end-effector pose. Among these, we choose the one where the elbow joint is concave down (like the pose in Figure~\ref{fig:figure_one}). This lets the robot lift boxes without having to flip its elbow orientation.


\subsection{Optimization Formulation} \label{optimization}
The backbone of \OptName{}'s trajectory generation and optimization is derived from DJ-GOMP~\cite{ichnowski2020djgomp}.
DJ-GOMP formulates a nonlinear optimization problem and solves it using sequential quadratic programming (SQP) to compute a jerk-limited, obstacle-avoiding, and grasp-optimized motion plan that is within the robot's dynamic limits. In DJ-GOMP, each SQP solve minimizes jerk (this empirically encourages smooth trajectories) given a fixed predicted trajectory duration. The minimum feasible duration is determined by repeated attempted solves with progressively shorter trajectory horizons. 

In \OptName{}, we extend the obstacle avoidance to the gripper, grasped box, and more complex collision environment. 
We summarize the \OptName{} optimization formulation here:
\begin{IEEEeqnarray*}{r?l/s}
    \argmin_{q_{[0..H]}} & \frac 12 \sum_{t=0}^{H-1} \lVert \dddot q_t \rVert^2_2 \\
    \text{s.t.}
    & q_t \in \mathcal{C}_\mathrm{free} & \small (1) \\ 
    & q_0 = q^d_0 ,\quad q_H = q^d_H & \small (2a), (2b) \\ 
    & \dot q_0 = \dot q_H = 0,\quad \ddot q_0 = \ddot q_H = 0 & \small (2c), (2d) \\ 
    & q_{t+1} = q_{t} + \dot q_{t} t_\mathrm{step} + \tfrac 12 \ddot q_{t} t_\mathrm{step}^2 + \tfrac 16 \dddot q_{t} t_\mathrm{step}^3 & \small (3a) \\ 
    & \dot q_{t+1} = \dot q_{t} + \ddot q_{t} t_\mathrm{step} + \tfrac 12 \dddot q_{t} t_\mathrm{step}^2  & \small (3b) \\ 
    & \ddot q_{t+1} = \ddot q_{t} + \dddot q_{t} t_\mathrm{step} & \small (3c) \\ 
    & q_t, \dot q_t, \ddot q_t, \dddot q_t \in Q, \dot Q, \ddot Q, \dddot Q & \small (4) \\ 
\end{IEEEeqnarray*}

$H \in \mathbb{Z}_+$ is the time horizon, or number of waypoints after the start, $t_\mathrm{step} \in \mathbb{R}_+$ is the time interval between waypoints, and the constraints with subscript $t$ are for all valid $t$.

Constraint (1) ensures a collision-free trajectory. Constraints (2a), (2b), (2c), and (2d) fix the trajectory to the desired endpoint configurations, velocities, and accelerations. Constraints (3a), (3b), and (3c) enforce consistent dynamics. Constraint (4) enforces actuation limits.


Following from prior work~\cite{ichnowski2020gomp,ichnowski2020djgomp}, the solver uses sequential quadratic programming (SQP).

For the first SQP solve, BOMP initializes the solver with a trajectory that is linearly interpolated in joint space between the start and goal configurations. For each subsequent solution (i.e., as $t_\mathrm{step}$ decreases), BOMP initializes the SQP solver with the trajectory from the previous solution.

Aside from the obstacle-avoidance constraints (1), which are non-convex, the remaining constraints are all linear in the decision variables. Therefore, only the obstacle-avoidance constraints must be linearized to form the problem as a locally valid quadratic program (QP). Following from prior work, we also soften this constraint by implementing it as an objective with a large initial weight of 10000. This results in better gradients for convergence when the initialization trajectory is in collision. 

Based on the success or failure of each QP solve iteration, we adjust the trust region (in which we expect our constraint linearization is approximately valid) and the weighting of the collision-free soft constraint.

Breaking from prior work, we change how we reduce the trajectory duration between SQP solves. Instead of reducing $H$, we keep $H$ fixed, and reduce $t_\mathrm{step}$ using
an empirically selected upper bound, 
and a binary search to find the lower bound. In experiments, we fix $H=16$ and use an initial $t_\mathrm{step}=160$\,ms based on a parameter sweep. We make this change primarily to reduce computation time. The QP solve time grows as $O(H^3)$ due to the number of optimization variables. With the more complicated collision environment in \AlgName{}, optimization became prohibitively slow with the large $H$ values (i.e., starting from $H=60$) used in prior work.
\label{collision_checking}
\subsection{Collision Checking}

During optimization, \OptName{} must repeatedly check for collisions between the robot, grasped box, and environment. Due to the long-nosed end-effector, the robot wrist is almost always outside of the bin, so we accelerate our collision checking by only checking for collisions between the long-nosed suction tool, the grasped box, and the environment. 

\begin{figure}
\centering
\includegraphics[width=0.48\textwidth]{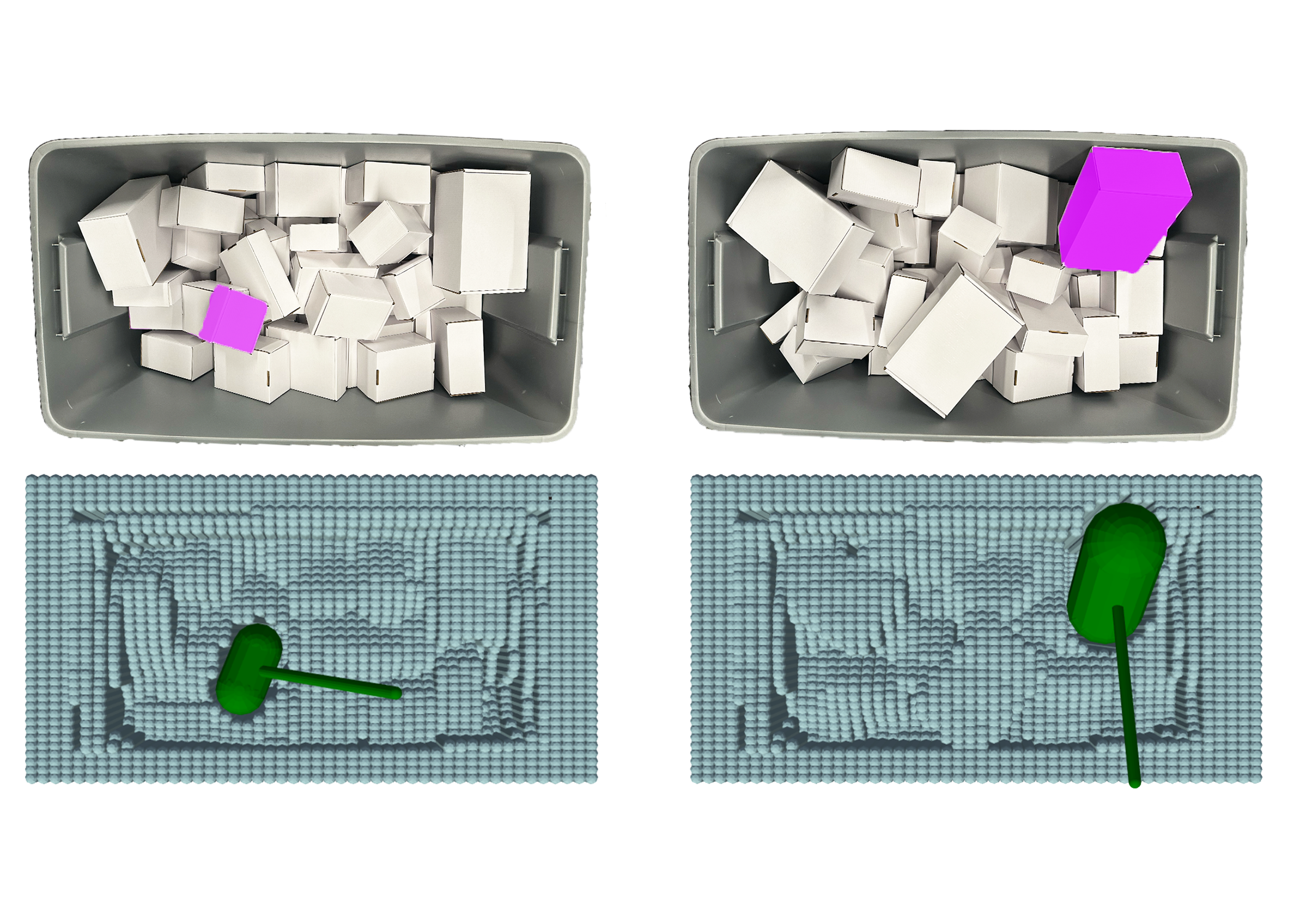}
\caption{\textbf{Example scenes and carved height maps}. 
The top images show bins with target boxes in pink. The bottom images show the capsule-modeled, padded, and carved height maps (blue) and the robot end-effector and grasped box capsule models (green). Note that the bin padding (solid height border) in the height maps prevents carving through the bin wall in the second scene.
}
\label{fig:scenes}
\end{figure}

\subsubsection{Collision Model}
Using $\mathbf{y}_\mathrm{D}$, we compute a height map collision environment, 
then downsample it by max-pooling to improve collision-checking efficiency. In experiments, we downsample to 30$\times$40.
%
We use capsules (cylinders with hemispherical caps) for collision checking because capsule-capsule distance checks have a fast, closed-form solution, roughly 10x faster than the box-box collision checking in Flexible Collision Library~\cite{pan2012fcl}. Capsules tightly bound the tall, thin height map columns at each pixel, and the tube-like suction tool.

We approximate each height map cell with a vertically-oriented capsule.
The cylindrical portion of the capsule extends from some ``world bottom'' $z_0$, outside the max reach of the robot, to the top of the height map column, $z_{ij}$. The set of all environment capsules is $\mathcal{Y}$. We define $\mathcal{R}$ as the set of capsules bounding the long-nosed suction tool and the grasped box. 
\begin{figure*}[t]
\centering
\includegraphics[width=1.0\textwidth]{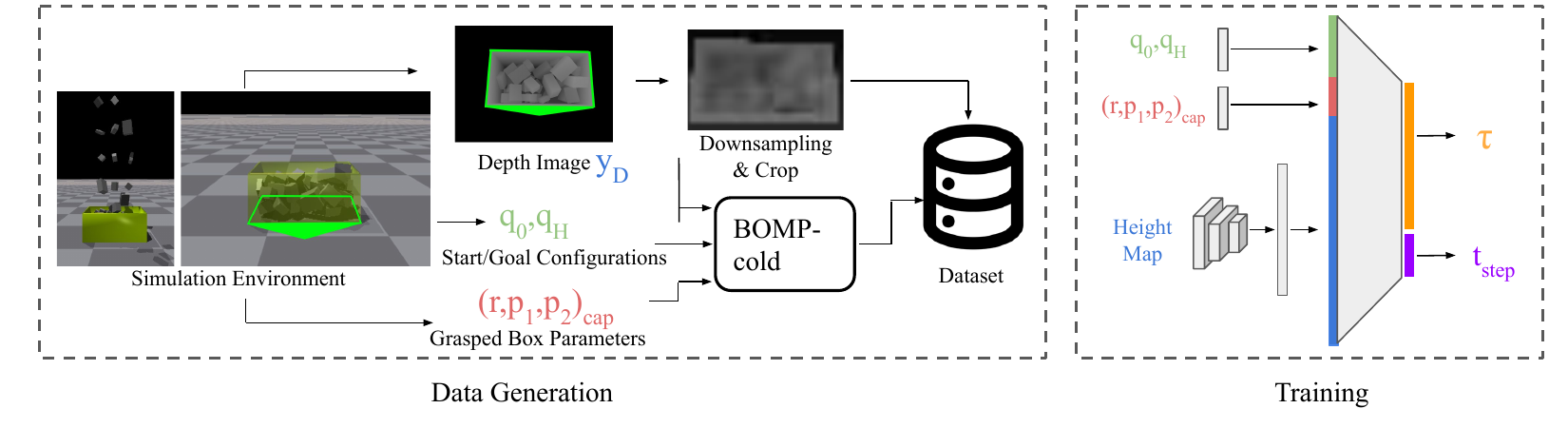}
\caption{ \textbf{Deep learning warm-start}. To create the training dataset (\textbf{left}), we drop boxes randomly in a simulated bin, then generate a downsampled depth map via max pooling. We select the topmost box, compute start and goal configurations and box dimensions, and pass them to cold-started \AlgName{} to generate a time-optimized trajectory. We save the optimizer inputs with the resulting trajectory in a dataset. The warm-start network (\textbf{right}) predicts a trajectory $\tau$ and segment duration $t_\mathrm{step}$ given a height map converted from a depth image, start and goal configurations, and the grasped box parameters.
}
\label{fig:pipeline}
\end{figure*}

\subsubsection{Collision Optimization}
During the optimization, we check for collisions between the robot end effector, grasped box, and the height map. For efficiency, we only check each capsule in $\mathcal{R}$ against the height map capsules that intersect its axis-aligned bounding box (ignoring z-coordinate). We define $d$ as the closest distance between 
capsules in $\mathcal{Y}$ and $\mathcal{R}$ at a given time along the trajectory. This is the minimum distance the capsule in $\mathcal{R}$ must move in any direction to no longer be in collision.
%

In the deep bin environment, it is easiest to avoid obstacles by going \emph{over} them (intuitively, the robot cannot pass through the walls of the bin so it must go out and over the top). We encode this intuition into the optimization by updating 
$d$ to instead be the minimum distance the capsule in $\mathcal{R}$ must move \emph{up} to no longer be in collision. To still allow trajectories around (i.e., not over) small obstacles, we make this update only when the $\mathbf{center}$ $\mathbf{axis}$ of the closest robot-fixed capsule intersects the closest environment capsule.

We follow CHOMP~\cite{ratliff2009chomp} in scaling $d$ by the robot's linear speed to disincentivize speeding through obstacles. We compute the robot's linear speed at the collision point by applying the manipulator Jacobian to the joint velocities at that point in the trajectory.
To account for collisions in between trajectory waypoints, we densely sample $d$ uniformly over each trajectory segment between $t$ and $t+1$ and sum the results to get $D_t$. (In experiments, we sample 50x over each segment based on a parameter sweep 
to find the minimum value that does not impact on success rate.) Defining $\tilde{D_t}$ as the closest distance using the last QP solve trajectory $\tilde\tau$, we define the linearized collision constraints as:
\begin{equation} \label{jacobian_calcs}
    \tilde{D_t}
    + \frac{\partial D_t}{\partial q_t}q_t 
    + \frac{\partial D_t}{\partial q_{t+1}}q_{t+1}
    + \frac{\partial D_t}{\partial\dot{q}_t}\dot{q}_t
    + \frac{\partial D_t}{\partial\dot{q}_{t+1}}\dot{q}_{t+1}
    > 0.
\end{equation}
We compute the above Jacobians using finite differences.

To limit failures from the optimizer stopping with slight (e.g., $<1$\,cm) collisions, we inflate the capsules in $\mathcal{R}$ (by 1\,cm), and add an equivalent acceptance tolerance. 
Thus, if the optimizer terminates within the tolerance, the trajectories are verifiably collision-free.




\subsubsection{Carving}
\label{carving}
To model the change of the grasped box from being a part of the obstacle environment to being attached to the robot end effector, we \emph{carve} out (remove) any capsule in $\mathcal{Y}$ in the same location as the capsule model for the grasped box. This prevents detecting non-existent collisions at $q_0$.
To avoid physically infeasible trajectories due to carving through bin walls, we artificially thicken the bin walls in the height map passed to the optimizer. 
%
Figure~\ref{fig:scenes} shows carving results. 

\subsection{Deep Learning Warm-start}
\label{sec:warmstart}
Even with optimized collision checking, computing fast trajectories in complex environments requires long compute times (roughly 30 seconds).
DJ-GOMP~\cite{ichnowski2020djgomp} showed that a neural network could speed up computation by outputting an initial guess for $\tau$ and $H$ (the trajectory and horizon) that could \emph{warm-start} the optimizer and allow it to converge faster.
We adapt this technique to reduce the computation time of \AlgName{} given an offline set of examples in simulation.
We train a neural network (Figure~\ref{fig:pipeline}) that predicts the optimal trajectory $\tau$ and the optimal segment duration $t_\mathrm{step}$ given start and end poses, radius and endpoints of the grasped capsule (in the end-effector frame), and the scene's height map. The neural network passes the height map through two convolutional layers to produce an embedding, which is concatenated with the start pose, end pose, radius, and endpoints of the grasped capsule. This concatenated vector is passed through an encoder to yield an initial "warm-start" trajectory and $t_\mathrm{step}$.

\subsubsection{Generating Training Data}
\label{sec:training} 
To generate training data, we use Isaac Gym~\cite{makoviychuk2021isaac} to simulate an environment with a rectangular bin that matches the real bin's dimensions and fill it with randomly sampled boxes with dimensions matching our real setup. A virtual depth camera captures an overhead depth image of the bin and converts it into a height map. To closely match the distributions of boxes through the picking cycle, we repeatedly select the topmost box using the same prompting method as in real (using the ground truth box shape instead of a SAM estimation) and remove it until the bin is sufficiently close to empty.
In experiments, the training set consists of 25,000 simulated scenes generated from a distribution of box sizes and box counts likely to appear at test time, along with corresponding height maps, grasped boxes, and time-optimized trajectories.

\subsubsection{Warm-starting}

Despite the widely varying bin contents between scenes, we are able to learn a useful warm-start by referencing the height map during training and inference. 

At test time, we prompt the network with the trajectory endpoints, grasped box parameters, and height map. We warm-start the optimizer with the network's predicted trajectory $\tau$ and segment duration $t_\mathrm{step}$.

We differentiate from DJ-GOMP, which does not include a grasped box or the height map in the inputs to the neural network. Furthermore, we predict $t_\mathrm{step}$ given a fixed $H$; DJ-GOMP did the opposite. This change simplifies the neural network architecture but requires interpolation to run on a robot's fixed-time-interval controllers.


\subsection{Speeding Up Computation}
We make several optimizations to speed up computation at test time. We parallelize the Jacobian calculations for the obstacle constraints (Equation~\ref{jacobian_calcs}) over 12 Intel Core i7-6850K CPUs at 3.60GHz. Like DJ-GOMP~\cite{ichnowski2020djgomp}, we solve only until a collision-free solution is found. Empirically, the additional solve time to find a fully optimal solution does not outweigh the additional computation time cost to generate it.
\section{Experiments} \label{experiments} 
We compare \OptName{} with two baselines: (1) an industry-standard Up-Over-Down motion that lifts the grasped box vertically, moves horizontally over the bin wall, then lowers; 
and (2) a parallelized asymptotically optimal sampling-based motion planner from Motion Planning Templates MPT~\cite{ichnowski2019mpt} with a subsequent time-parameterization step~\cite{kunz2012time}.

We compare the algorithms on compute time (time to compute a trajectory given a suction point and grasped box dimensions), execution time, total time (compute time $+$ execution time), and collision-free trajectory generation rate (``success rate''). In physical experiments, we also report ``execution success rate'' which considers the percent of all experiments (including those which failed in generation) where the robot successfully transports the box from start to goal. 


Due to reachability constraints associated with our robot, suction tool, and environment, a fully vertical extraction is often not possible. Thus, the Up-Over-Down baseline performs a two-stage ``up'' motion. The first stage moves the end-effector vertically as far as possible; the second stage moves the box vertically as far as possible while pitching $22.5^{\circ}$ back to enable the box to be raised higher when a top-down grasp is no longer kinematically feasible. Due to the deep bin (compared to the size of the robot), from most locations in the bin, it is kinematically infeasible for the robot to lift the box higher than the rim of the bin with a top-down grasp.

We use MPT's parallelized RRT* (PRRT*~\cite{ichnowski2014prrt}), as it is a fast asymptotically optimal planner.
We configure it to optimize for minimum path length in joint space. 
We run MPT for 1 second (MPT-1), which is similar to \OptName{}'s generation time, and MPT for 10 seconds (MPT-10), which has more time to find and optimize a solution. While PRRT* will eventually find a solution if one exists, it may fail with insufficient planning time.
Since the MPT baselines use random sampling, we execute them three times each per scene and average the metrics. We only include successfully computed trajectories in the calculation for the time-based metrics. 

We also perform ablations to assess the impact of warm-starting and training the neural network on the height map.
We isolate the deep-learning warm-start by including cold-started optimization with an empirically selected heuristic $t_\mathrm{step}$ of 160\,ms, labeled \AlgName{}-t160ms in Table\,\ref{tab:sim-result} and Table\,\ref{tab:sim-phys}.
We also consider the impact of the neural network knowing the height map in \AlgName{}-NH (``no heightmap''). For this ablation, we perform neural network inference \emph{without} an input height map.
As in DJ-GOMP~\cite{ichnowski2020djgomp}, since additional SQP solves empirically take more time to compute than they save in execution time, we stop \AlgName{}, \AlgName{}-t160ms, and \AlgName{}-NH after they first find a feasible trajectory.
To compare against optimal execution times, we also show results from the cold-started optimization when run to optimal convergence (labeled \AlgName{}-cold).

For both simulated and physical scenes, we model a UR5 robot reaching into a deep bin (dimensions 1.06$\times$0.562$\times$0.46\,m$^3$) full of boxes. To allow the robot to reach and manipulate boxes deep in the bin, we equip it with the ``bluction'' tool from  Huang, et al.~\cite{huang2022bluction} (blade and camera attachments removed).

We fill the bin with cardboard boxes of assorted sizes. We use between 5 and 15 boxes of each type to fill the bin. For our experiments, we use boxes that are 4$\times$4$\times$2\,in$^3$, 6$\times$4$\times$3\,in$^3$, 7$\times$5$\times$2\,in$^3$, and 9$\times$6$\times$3\,in$^3$ because:
\begin{enumerate}
\item They represent a range of aspect ratios (which directly affects the fidelity of the capsule model).
\item The 6$\times$4$\times$3\,in$^3$ and 9$\times$6$\times$3\,in$^3$ boxes have the 6$\times$3\,in$^2$ face in common. We demonstrate that our pipeline is robust to this.
\end{enumerate}


\begin{figure}[t]
\centering
\includegraphics[width=0.47\textwidth]{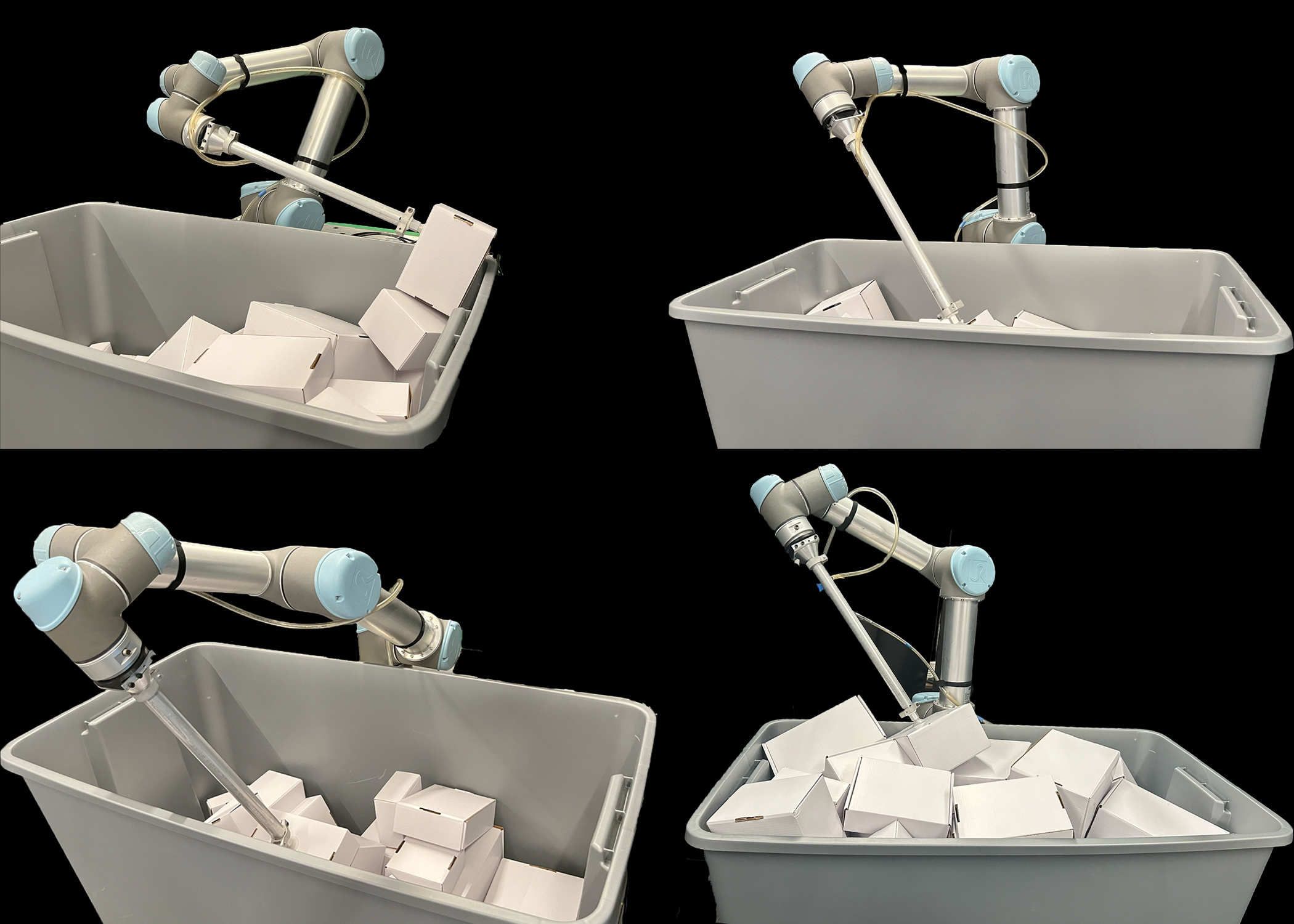}
\caption{\textbf{Challenging grasp poses}. In physical experiments, we observe that using the long suction tool to grasp arbitrarily oriented boxes sometimes results in challenging grasp poses such as the ones pictured here. While the industry-standard Up-Over-Down method fails in these cases, BOMP is able to generate fast, jerk-limited, collision-free trajectories.
}
\label{fig:challenging_scenes}
\end{figure}

\begin{table*}[h]
\vspace{6pt}
\begin{adjustbox}{center}
\begin{tabular}{r | r | r | r | r}
    \toprule
    \textbf{Algorithm}&\textbf{Success Rate}&\textbf{Exec. Time(s)}&\textbf{Compute Time(s)}& \textbf{Total Time(s)} \\
    \hline 
    \hline
    MPT-1 & 69.44\% & 1.941$\pm$0.776 & 1.833$\pm$0.044 & 3.774$\pm$0.786 \\
    MPT-10 & \textbf{84.38\%} & 1.941$\pm$1.023 & 10.872$\pm$0.049 & 12.813$\pm$1.042 \\
    Up-Over-Down & \textcolor[HTML]{ff0000}{16.67\%} & 1.689$\pm$0.225 & \textbf{0.837$\pm$0.158} & \textbf{2.525$\pm$0.293} \\
    \hline
    \rule{0pt}{2.5ex} \AlgName{}-NH & 80.21\% & 1.209$\pm$0.146 & 1.664$\pm$2.498 & 2.872$\pm$2.496 \\
    \AlgName{}-cold & \textbf{84.38\%} & \textbf{0.982$\pm$0.352} & 33.696$\pm$20.067 & 34.678$\pm$20.002 \\
    \AlgName{}-t160ms & \textbf{84.38\%} & 2.560$\pm$0.000 & 1.661$\pm$1.144 & 4.221$\pm$1.144 \\
    \AlgName{} & 78.13\% & 1.080$\pm$0.142 & 1.481$\pm$1.639 & 2.561$\pm$1.649 \\
    \bottomrule
\end{tabular}

\end{adjustbox}
\caption{\textbf{Simulated experiments results.} In 96 feasible simulated environments, we execute 3 trials for the sampling-based methods (MPT-1 and MPT-10) and 1 trial for the deterministic methods. MPT-10 and the cold-started \AlgName{} ablations have the highest success rate. With the warm-start, \AlgName{} achieves a similarly high success rate while also achieving the fastest total time aside from the unreliable Up-Over-Down. }
\label{tab:sim-result}
\end{table*}
\begin{table*}[h] 
\vspace{6pt}
\begin{adjustbox}{center}
\begin{tabular}{r | r | r | r |r | r}
    \toprule
    \rule{0pt}{2.5ex} \multirow{2}{*}{\textbf{Algorithm}}
    & \textbf{Generation}& \textbf{Execution}  & 
    \multirow{2}{*}{\textbf{Exec. Time(s)}}  & 
    \multirow{2}{*}{\textbf{Compute Time(s)}} &
    \multirow{2}{*}{\textbf{Total Time(s)}}\\
    & \textbf{Success Rate}& \textbf{Success Rate} 
    & \\
    \hline
    \hline
    \rule{0pt}{2.5ex}MPT-1 & 68.89\% & 53.33\% & 2.448$\pm$1.319 & 1.747$\pm$0.137 & 4.194$\pm$1.334 \\
    MPT-10 & \textbf{93.33\%} & 66.67\% & 2.210$\pm$1.269 & 10.792$\pm$0.140 & 13.001$\pm$1.300 \\
    Up-Over-Down & \textcolor[HTML]{ff0000}{6.67\%} & \textcolor[HTML]{ff0000}{6.67\%} & 1.432$\pm$0.000 & \textbf{0.710$\pm$0.000} & \textbf{2.142$\pm$0.000} \\
    \hline
    \rule{0pt}{2.5ex}\AlgName{}-cold & 86.67\% & 73.33\% & \textbf{0.884$\pm$0.279} & 34.074$\pm$14.616 & 34.958$\pm$14.590 \\
    \AlgName{} & 86.67\% & \textbf{80.00\%} & 1.035$\pm$0.168 & 1.583$\pm$1.075 & 2.617$\pm$1.067 \\
    \bottomrule
\end{tabular}

\end{adjustbox}
\caption{\textbf{Physical experiments results.} In 15 physical environments, we perform 3 trials for the non-deterministic MPT-1 and MPT-10 and 1 trial for the deterministic methods. \AlgName{} successfully executes the most trajectories and achieves the fastest total time (except Up-Over-Down, which only successfully executes 1 trajectory). MPT-1 and MPT-10 generate the most theoretically feasible trajectories but their jerky, non-smooth trajectories result in several dropped boxes and automatic protective stops during physical execution.}
\label{tab:sim-phys}
\end{table*}

\subsection{Simulated Experiments}
We generate 114 simulated environments using the same process used to generate neural network training data (described in Section~\ref{sec:training} and shown in Figure~\ref{fig:pipeline}). However, these scenes were not previously seen by the neural network.

We exclude generated environments where the initial grasp is unreachable or results in an IK solution that is in collision. Results are shown over the remaining 96 feasible scenes.


For each environment, we capture a simulated depth image, convert it into a height map, and select the topmost box to be removed. In simulation, rather than using SAM~\cite{kirillov2023segany} to determine box pose and size (Section \ref{sec:sam}), we use ground truth box pose and size to inform the solvers of the grasped box. We define the goal at a joint configuration where all in-distribution grasped boxes can be held without collision.
For a fair comparison, we use the capsule model (Section~\ref{collision_checking}) to evaluate collisions for all planners and we carve the height maps as in Figure~\ref{fig:scenes} before solving.

We then use \AlgName{}, 
baselines, and ablations
to plan trajectories that carry the selected box from its start point to the endpoint. Table~\ref{tab:sim-result} displays computation success rate, compute time, trajectory execution time, and total time (compute time $+$ execution time) from these experiments.

We find that MPT-10 and the cold-started \AlgName{} ablations have the highest success rate. 
With the warm-start, \AlgName{} strikes a balance between fast computation and fast collision-free trajectories. It achieves a similarly high success rate to MPT-10, a fast computation time close to \AlgName{}-160ms, and a fast execution time close to \AlgName{}-cold.

Up-Over-Down achieves a slightly faster total time than \AlgName{} due to its negligible computation time, but Up-Over-Down's success rate is very low. This is primarily due to the kinematic difficulty of safely vertically extracting \emph{arbitrarily oriented} boxes, where the suction normal is generally not aligned to gravity (Figure~\ref{fig:challenging_scenes}).

\AlgName{}-NH, without a height map of the environment, tends to predict trajectories closer to ``average'' than \AlgName{}. This means larger (less optimized) values of $t_\mathrm{step}$ and trajectories that are on average further from optimal. This explains the increase in trajectory execution time and success rate. The higher predicted $t_\mathrm{step}$ also means that it is less often below the optimal value (i.e., infeasible). This results in the slightly higher success rate. On average, \AlgName{}-NH's computation time is 0.183\,s higher than \AlgName{}, suggesting that the network's knowledge of the obstacles through the height map generates trajectories that are more favorable for warm-starting.

In 93\% of failure cases for \AlgName{}-cold and \AlgName{}-t160ms, the target box starts in contact with the bin wall. In these cases, the collision environment near the grasped box is particularly dense. The warm-started \AlgName{} and \AlgName{}-NH also occasionally under-predict the optimal segment duration $t_\mathrm{step}$, resulting in their slightly lower success rates.


\subsection{Physical Experiments}
\label{sec:phys-exps}
We perform similar experiments using 15 physical setups (see examples of the physical environment in Figure~\ref{fig:scenes}). We use an overhead Intel RealSense D455 camera to capture depth images for collision avoidance and RGB images for segmentation. These images are 480$\times$640 pixels, but we downsample the height map to 30$\times$40 for collision checking (Section~\ref{collision_checking}). We detect and select grasped boxes using the full grasp selection pipeline (Section \ref{sec:sam}). 

As in simulated experiments, we consider only scenes with reachable, collision-free grasp poses.
For the non-deterministic MPT-1 and MPT-10, we average metrics across three trials.
The results of these experiments are in Table~\ref{tab:sim-phys}.

In physical experiments, \AlgName{} successfully executes the most trajectories and achieves the fastest total time (compute time $+$ execution time) other than Up-Over-Down, which only successfully executes 1 trajectory. While MPT-10 generates the most theoretically feasible trajectories, these often result in dropped boxes or automatic protective stops because its trajectories are not jerk-limited. This particularly happens when the trajectory turns sharp, high-jerk ``corners'' and at the trajectory endpoint. \AlgName{} computes a smooth trajectory that better avoids protective stops and drops.

At execution time, the failure cases for \AlgName{} and \AlgName{}-cold are primarily caused by approximation error during carving (Section~\ref{carving}). The capsule model overestimates the actual box volume, so it may intersect other obstacles. The intersected parts of the obstacles are removed from the collision model by the carving process. When the resulting trajectories are physically executed, these trimmed obstacles may dislodge the box from the robot gripper.

Up-Over-Down has an extremely low success rate in this environment due to the complex grasp poses (not simply top down).
Figure \ref{fig:challenging_scenes} shows some example scenes with complex grasp poses from which \AlgName{} is able to compute a solution, but Up-Over-Down is not.

\section{Discussion}
We present \AlgName{}, an optimization-based motion planner integrated into an end-to-end bin picking pipeline.
We integrate a grasp and target box identification method using the Segment Anything Model~\cite{kirillov2023segany}, and we warm-start the motion planning optimization on the output of a deep neural network trained to consider obstacles in the form of a height map. 
We train the network offline on simulated data, then use the trained network online to provide an initial guess of the trajectory and its duration. We use this output to warm-start the optimizer and speed up its convergence.

In 15 experiments in real bin scenes and 96 experiments in simulation, \AlgName{} outperforms heuristic and sampling-based baselines in execution time by up to 36\% and 58\% respectively, while generating jerk-limited trajectories. \AlgName{} also achieves the fastest total time (compute time $+$ execution time) among methods with comparable success rate.

In future work, we will address several limitations. We plan to extend beyond known boxes to general unseen grasped objects by using SAM~\cite{kirillov2023segany} alongside a grasp planner such as Dex-Net 3.0~\cite{mahler2018dexnet3}. We will also speed up SAM prompting by using a smaller model fine-tuned for the bin.

We also plan to improve the fidelity of the capsule modeling. The capsule model overestimates the extents of the grasped box, so it sometimes carves into nearby obstacles. As a result, the motion planner may find a solution that collides in real since not all of the true collision environment is present in the carved height map it uses to plan.

Within these limitations, though, \AlgName{} significantly outperforms baselines in speed while maintaining a comparable or superior success rate.


\section*{Acknowledgments}
This research was performed at the AUTOLAB at UC Berkeley in affiliation with the Berkeley AI Research (BAIR) Lab. The authors were supported in part by donations from Toyota Research
Institute, Bosch, Google, Siemens, and Autodesk.

\bibliographystyle{IEEEtran}
\bibliography{references}

\begin{thebibliography}{10}
\providecommand{\url}[1]{#1}
\csname url@samestyle\endcsname
\providecommand{\newblock}{\relax}
\providecommand{\bibinfo}[2]{#2}
\providecommand{\BIBentrySTDinterwordspacing}{\spaceskip=0pt\relax}
\providecommand{\BIBentryALTinterwordstretchfactor}{4}
\providecommand{\BIBentryALTinterwordspacing}{\spaceskip=\fontdimen2\font plus
\BIBentryALTinterwordstretchfactor\fontdimen3\font minus \fontdimen4\font\relax}
\providecommand{\BIBforeignlanguage}[2]{{%
\expandafter\ifx\csname l@#1\endcsname\relax
\typeout{** WARNING: IEEEtran.bst: No hyphenation pattern has been}%
\typeout{** loaded for the language `#1'. Using the pattern for}%
\typeout{** the default language instead.}%
\else
\language=\csname l@#1\endcsname
\fi
#2}}
\providecommand{\BIBdecl}{\relax}
\BIBdecl

\bibitem{huang2022bluction}
H.~Huang, M.~Danielczuk, C.~M. Kim, L.~Fu, Z.~Tam, J.~Ichnowski, A.~Angelova, B.~Ichter, and K.~Goldberg, ``Mechanical search on shelves using a novel “bluction” tool,'' in \emph{2022 IEEE ICRA}.

\bibitem{ichnowski2020gomp}
J.~Ichnowski, M.~Danielczuk, J.~Xu, V.~Satish, and K.~Goldberg, ``Gomp: Grasp-optimized motion planning for bin picking,'' in \emph{2020 IEEE ICRA}.

\bibitem{ichnowski2020djgomp}
J.~Ichnowski, Y.~Avigal, V.~Satish, and K.~Goldberg, ``Deep learning can accelerate grasp-optimized motion planning,'' \emph{Science Robotics}, vol.~5, no.~48, 2020.

\bibitem{kunz2012time}
T.~Kunz and M.~Stilman, ``Time-optimal trajectory generation for path following with bounded acceleration and velocity,'' \emph{Robotics: Science and Systems VIII}, pp. 1--8, 2012.

\bibitem{ichnowski2019mpt}
J.~Ichnowski and R.~Alterovitz, ``Motion planning templates: A motion planning framework for robots with low-power cpus,'' in \emph{2019 IEEE ICRA}.

\bibitem{ratliff2009chomp}
N.~Ratliff, M.~Zucker, J.~A. Bagnell, and S.~Srinivasa, ``Chomp: Gradient optimization techniques for efficient motion planning,'' in \emph{2009 IEEE ICRA}.

\bibitem{kalakrishnan2011stomp}
M.~Kalakrishnan, S.~Chitta, E.~Theodorou, P.~Pastor, and S.~Schaal, ``Stomp: Stochastic trajectory optimization for motion planning,'' in \emph{2011 IEEE ICRA}.

\bibitem{schulman2013finding}
J.~Schulman, J.~Ho, A.~X. Lee, I.~Awwal, H.~Bradlow, and P.~Abbeel, ``Finding locally optimal, collision-free trajectories with sequential convex optimization.'' in \emph{Robotics: science and systems}, vol.~9, no.~1.\hskip 1em plus 0.5em minus 0.4em\relax Citeseer, 2013, pp. 1--10.

\bibitem{kuntz2017fast}
A.~Kuntz, C.~Bowen, and R.~Alterovitz, ``Fast anytime motion planning in point clouds by interleaving sampling and interior point optimization,'' \emph{ISRR, 2017}, 2017.

\bibitem{marcucci2022motion}
T.~Marcucci, M.~Petersen, D.~von Wrangel, and R.~Tedrake, ``Motion planning around obstacles with convex optimization,'' \emph{Science Robotics}, vol.~8, no.~84, 2023.

\bibitem{natarajan2024implicit}
R.~Natarajan, C.~Liu, H.~Choset, and M.~Likhachev, ``Implicit graph search for planning on graphs of convex sets,'' \emph{Robotics: Science and Systems (RSS)}, 2024.

\bibitem{ichnowski2020deep}
J.~Ichnowski, Y.~Avigal, V.~Satish, and K.~Goldberg, ``Deep learning can accelerate grasp-optimized motion planning,'' \emph{Science Robotics}, vol.~5, no.~48, p. eabd7710, 2020.

\bibitem{hinton2015distilling}
G.~Hinton, O.~Vinyals, and J.~Dean, ``Distilling the knowledge in a neural network,'' \emph{arXiv:1503.02531}, 2015.

\bibitem{bucilua2006model}
C.~Buciluǎ, R.~Caruana, and A.~Niculescu-Mizil, ``Model compression,'' in \emph{Proceedings of the 12th ACM SIGKDD International Conference on Knowledge Discovery and Data Mining}.\hskip 1em plus 0.5em minus 0.4em\relax ACM, 2006, pp. 535--541.

\bibitem{mordatch2014combining}
I.~Mordatch and E.~Todorov, ``Combining the benefits of function approximation and trajectory optimization.'' in \emph{Robotics: Science and Systems}, vol.~4, 2014.

\bibitem{taylor2009transfer}
M.~E. Taylor and P.~Stone, ``Transfer learning for reinforcement learning domains: A survey,'' \emph{Journal of Machine Learning Research}, vol.~10, no. Jul, pp. 1633--1685, 2009.

\bibitem{lembono2020learning}
T.~S. Lembono, C.~Mastalli, P.~Fernbach, N.~Mansard, and S.~Calinon, ``Learning how to walk: Warm-starting optimal control solver with memory of motion,'' in \emph{2020 IEEE ICRA}.

\bibitem{mansard2018using}
N.~Mansard, A.~DelPrete, M.~Geisert, S.~Tonneau, and O.~Stasse, ``Using a memory of motion to efficiently warm-start a nonlinear predictive controller,'' in \emph{2018 IEEE ICRA}.

\bibitem{ye2017guided}
G.~Ye and R.~Alterovitz, ``Demonstration-guided motion planning,'' \emph{ISRR, 2017}.

\bibitem{abbeel2008apprenticeship}
P.~Abbeel, D.~Dolgov, A.~Y. Ng, and S.~Thrun, ``Apprenticeship learning for motion planning with application to parking lot navigation,'' in \emph{2008 IEEE/RSJ IROS}.

\bibitem{rana2017towards}
M.~A. Rana, M.~Mukadam, S.~R. Ahmadzadeh, S.~Chernova, and B.~Boots, ``Towards robust skill generalization: Unifying learning from demonstration and motion planning,'' in \emph{2017 CoRL}.

\bibitem{huang2024neural}
Z.~Huang, H.~Chen, J.~Pohovey, and K.~Driggs-Campbell, ``Neural informed rrt*: Learning-based path planning with point cloud state representations under admissible ellipsoidal constraints,'' in \emph{2024 IEEE ICRA}.

\bibitem{das2024motion}
D.~Das, Y.~Lu, E.~Plaku, and X.~Xiao, ``Motion memory: Leveraging past experiences to accelerate future motion planning,'' in \emph{2024 IEEE ICRA}.

\bibitem{ichter2018learning}
B.~Ichter, J.~Harrison, and M.~Pavone, ``Learning sampling distributions for robot motion planning,'' in \emph{2018 IEEE ICRA}.

\bibitem{zucker2008adaptive}
M.~Zucker, J.~Kuffner, and J.~A. Bagnell, ``Adaptive workspace biasing for sampling based planners,'' in \emph{2008 IEEE ICRA}.

\bibitem{qureshi2019motion}
A.~H. Qureshi, A.~Simeonov, M.~J. Bency, and M.~C. Yip, ``Motion planning networks,'' in \emph{2019 IEEE ICRA}.

\bibitem{schmitt2017sampling}
P.~S. Schmitt, W.~Neubauer, W.~Feiten, K.~M. Wurm, G.~V. Wichert, and W.~Burgard, ``Optimal, sampling-based manipulation planning,'' in \emph{2017 IEEE ICRA}.

\bibitem{ichnowski2014prrt}
J.~Ichnowski and R.~Alterovitz, ``Scalable multicore motion planning using lock-free concurrency,'' \emph{IEEE Transactions on Robotics}, vol.~30, no.~5, pp. 1123--1136, 2014.

\bibitem{coleman2014thunder}
D.~Coleman, I.~A. Şucan, M.~Moll, K.~Okada, and N.~Correll, ``Experience-based planning with sparse roadmap spanners,'' in \emph{2015 IEEE ICRA}.

\bibitem{berenson2012lightning}
D.~Berenson, P.~Abbeel, and K.~Goldberg, ``A robot path planning framework that learns from experience,'' in \emph{2012 IEEE ICRA}.

\bibitem{kirillov2023segany}
A.~Kirillov, E.~Mintun, N.~Ravi, H.~Mao, C.~Rolland, L.~Gustafson, T.~Xiao, S.~Whitehead, A.~C. Berg, W.-Y. Lo, P.~Doll{\'a}r, and R.~Girshick, ``Segment anything,'' \emph{arXiv:2304.02643}, 2023.

\bibitem{mahler2017dex}
J.~Mahler, J.~Liang, S.~Niyaz, M.~Laskey, R.~Doan, X.~Liu, J.~A. Ojea, and K.~Goldberg, ``Dex-net 2.0: Deep learning to plan robust grasps with synthetic point clouds and analytic grasp metrics,'' \emph{Robotics: Science and Systems (RSS)}, 2017.

\bibitem{zhao2023fast}
X.~Zhao, W.~Ding, Y.~An, Y.~Du, T.~Yu, M.~Li, M.~Tang, and J.~Wang, ``Fast segment anything,'' \emph{arXiv:2306.12156}, 2023.

\bibitem{fischler1981ransac}
M.~A. Fischler and R.~C. Bolles, ``Random sample consensus: a paradigm for model fitting with applications to image analysis and automated cartography,'' \emph{Communications of the Association for Computing Machinery (ACM)}, vol.~24, no.~6, p. 381–395, June 1981.

\bibitem{pan2012fcl}
J.~Pan, S.~Chitta, and D.~Manocha, ``Fcl: A general purpose library for collision and proximity queries,'' in \emph{2012 IEEE ICRA}.

\bibitem{makoviychuk2021isaac}
V.~Makoviychuk, L.~Wawrzyniak, Y.~Guo, M.~Lu, K.~Storey, M.~Macklin, D.~Hoeller, N.~Rudin, A.~Allshire, A.~Handa, and G.~State, ``Isaac gym: High performance gpu-based physics simulation for robot learning,'' \emph{arXiv:2108.10470}, 2021.

\bibitem{mahler2018dexnet3}
J.~Mahler, M.~Matl, X.~Liu, A.~Li, D.~Gealy, and K.~Goldberg, ``Dex-net 3.0: Computing robust vacuum suction grasp targets in point clouds using a new analytic model and deep learning,'' in \emph{2018 IEEE ICRA}.

\end{thebibliography}

\end{document}